\documentclass{article}

\usepackage[final]{neurips_2022}

\usepackage[utf8]{inputenc} 
\usepackage[T1]{fontenc}    
\usepackage{hyperref}       
\usepackage{url}            
\usepackage{booktabs}       
\usepackage{amsfonts}       
\usepackage{nicefrac}       
\usepackage{microtype}      
\usepackage{xcolor}         
\usepackage{todonotes}
\usepackage{amsmath}
\usepackage{caption}
\title{Data-Driven Plasticity Modeling via Acoustic Profiling}

\author{%
  Khalid El-Awady\thanks{This report is in collaboration with Drs. M. Omar and J. El-Awady of the Mechanical Engineering Dept. at John Hopkins University. They provided the experimental data and the motivation for the work, set the direction of exploration, and reviewed and evaluated the results. The author is very appreciative of all their efforts.} \\
  (\texttt{kae@stanford.edu}) \\
}

\begin{document}

\maketitle


\vspace{-0.4in}
\section{Problem Statement and Approach}

In crystalline metals, plastic deformation proceeds through abrupt, localized events rather than smooth flow. Permanent deformation is carried by dislocations—line defects in the crystal lattice—that move under applied stress but are often temporarily pinned by obstacles such as other dislocations, impurities, or microstructural features. When the local stress exceeds a critical threshold, groups of dislocations suddenly depin and move collectively over short distances, producing rapid redistributions of internal stress. These sudden events generate broadband elastic waves that propagate through the solid and are detected as acoustic emissions (AEs), a mechanism established in early foundational studies of acoustic emission during plastic deformation \cite{Scruby87}. 

AE signal processing has progressed from threshold-based event counting and simple spectral analysis to sophisticated time–frequency and statistical methods that reveal avalanche dynamics \cite{Terchi01}, \cite{berta24}. But the underlying mechanisms for AE events is still not fully understood and it is thought that AE events comprise a number of underlying constituent events operating at different scales from the atomic to the bulk material level. Some of these might be encoded in the frequency spectrum. 

Thus the field remains ripe for modern nonstationary modeling, representation learning, and sequence-based signal analysis. We endeavor to extend current analysis of AE waveforms from retrospective characterization toward temporal modeling and forecasting, breaking down AE into a series of nonstationary banded signals whose early evolution encodes information about ongoing material deformation mechanics. Our approach includes:
\begin{enumerate}

    \item Detection of AE events localized to specific frequency bands of concentrated energy using wavelets in a fashion akin to seismic wave detection. 

    \item Physics-based analysis of the waveforms to validate the meaningfulness of the detected events and provide a plausible justification and interpretation of the results.

    \item Extraction of a labeled dataset of AE events and non-events that can be used to further identify and study features of AE events that are useful in understanding the underlying mechanics. 

    \item Exploration of signal classifiers that could be used in further detection and understanding of the underlying dislocation dynamics 
    
\end{enumerate}

\section{Experimental Setup and Raw Data}
\begin{figure}[!ht]
\begin{center}
    \includegraphics[width=1.1\linewidth]{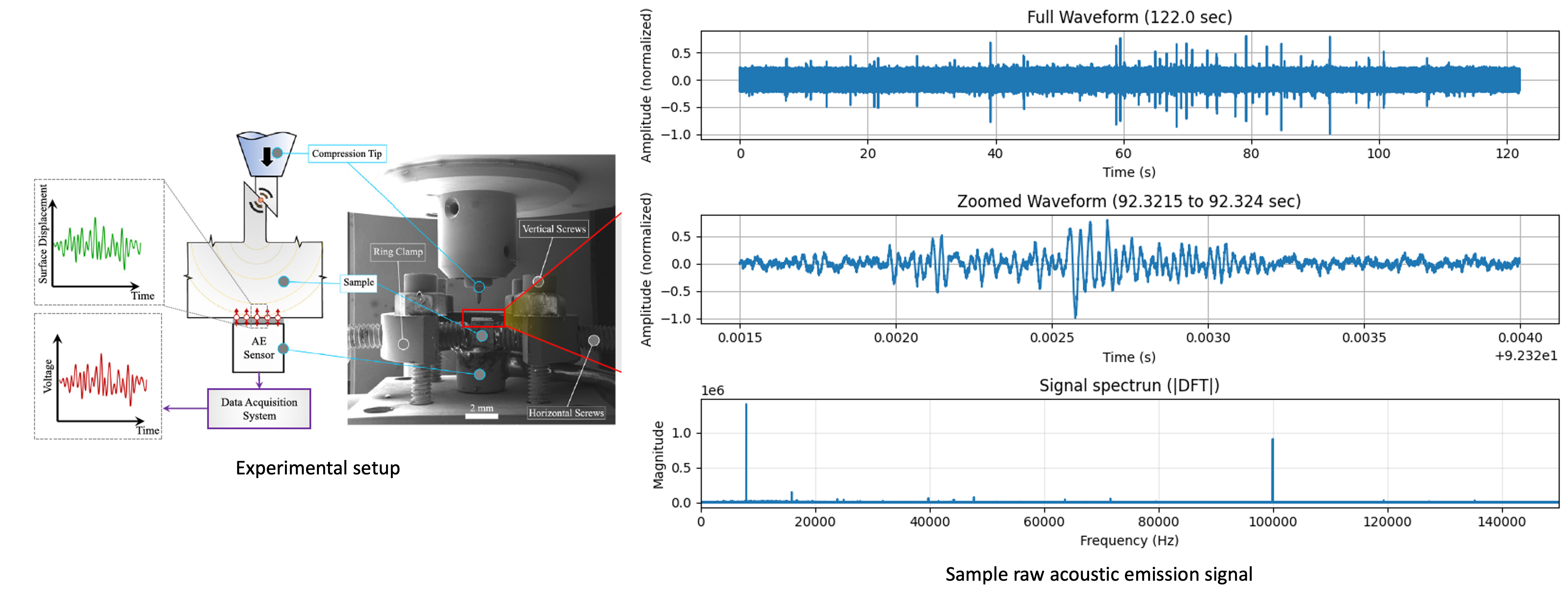}
    \caption{Experimental setup for the original data acquisition (left) and acquired acoustic emission plot and its magnitude spectrum (right).} 
    \vspace{-0.2in}
    \label{fig:experiment}
\end{center}    
\end{figure}
This work builds on the experiments of Dr. J. El-Awady and Dr. M. Omar of John Hopkins University who have collected an AE dataset for the research published here: \cite{Omar25-1}, \cite{Omar25-2}. Figure (\ref{fig:experiment}) shows the experimental setup used and a sample AE waveform. The authors load a small micropillar of Nickel ($3.3 \times 3.3 \times 9.0$ mm$^2$) with compressive stress that increases over time. The pillar undergoes strain, moving from elastic deformation through plastic deformation. At various times in this process the material experiences sudden dislocations and in the process releases energy in the form of acoustic emissions (AEs) that resemble seismic events (\cite{Ispanovity75, Baro13}). The AEs are detected via a piezo-electric device attached to the pillar. The right side of figure (\ref{fig:experiment}) shows the acquired AE signal over the approximately 2 min duration of the experiment.  The signal is acquired at a 2MHz sampling rate then normalized to a $\pm 1$ amplitude. The signal suffers from significant background and measurement noise as seen in the early stages of the signal when deformation is minimal and no dislocations should be present.

In the authors' original work the occurrence of AE events is detected via amplitude thresholding. The noise magnitude level is estimated from the early dislocation-free portion of the signal and AE events are then identified based on the signal exceeding the background noise. While this approach does capture the major AE events, it may be missing ones of smaller magnitude, hampering a complete understanding of the evolution of the plastic deformation.

\section{Wavelet-based AE Event Detection}
\begin{figure}[!ht]
\begin{center}
    \includegraphics[width=0.7\linewidth]{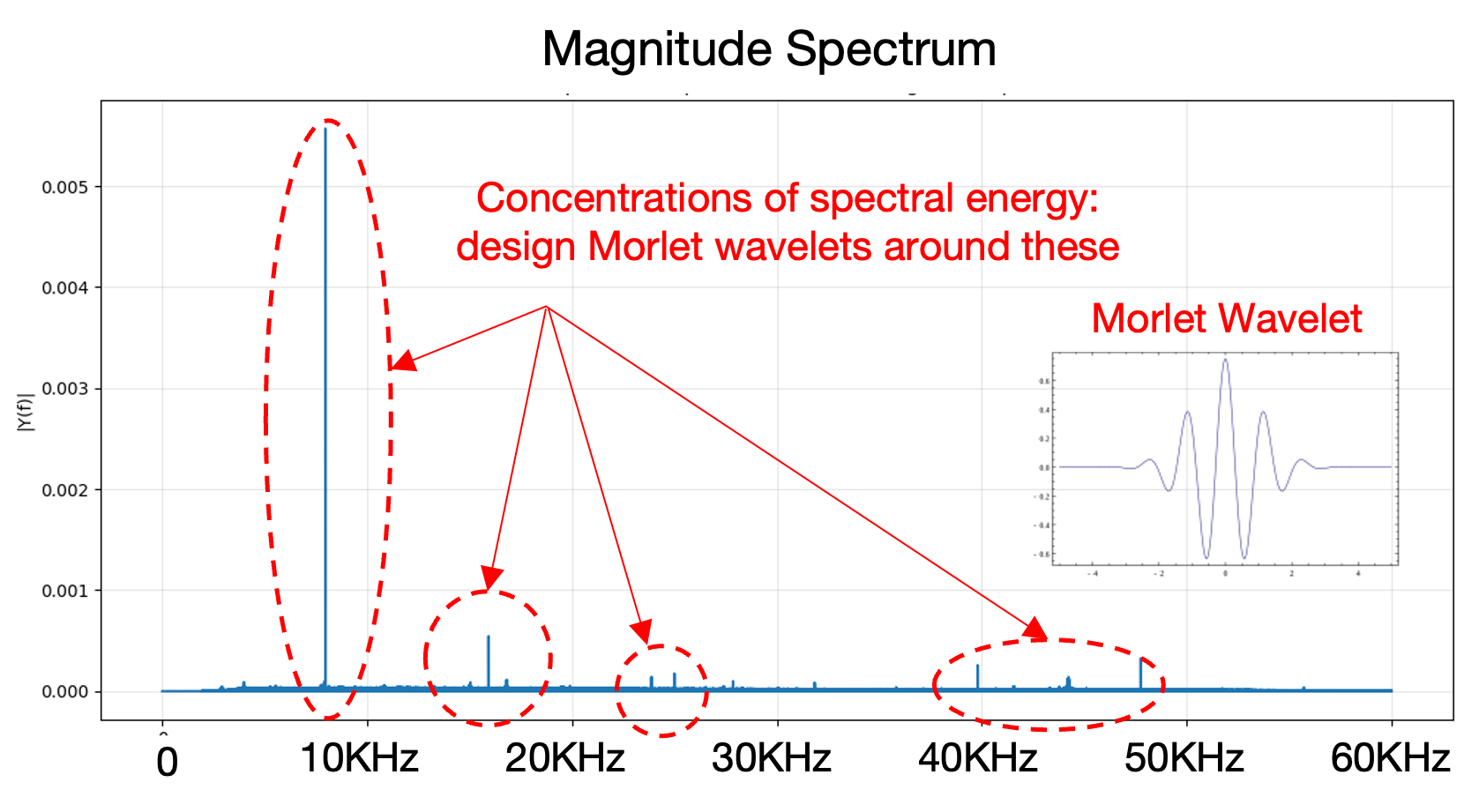}
    \caption{Magnitude spectrum of the signal after apply a band-pass filter in the frequency range 2KHz to 60 KHz. The spectrum exhibits some prominent peaks after filtering in the vicinities of 8KHz, 16KHz, 25 KHz, and 44 KHz.} 
    \vspace{-0.2in}
    \label{fig:p36_filt_2KHz_60KHz}
\end{center}    
\end{figure}
We explore an alternative approach to AE event detection based on Morlet wavelets (\cite{Jiang22}). First we note that \cite{Omar25-1} indicate that the signal spectrum of interest based on the problem physics lies below 60 KHz. We apply a zero-phase bandpass filter to the signal in the range of 2KHz to 60 KHz to reduce the impact of the noise. The filtered magnitude spectrum is shown in figure (\ref{fig:p36_filt_2KHz_60KHz}). The spectrum shows concentrations of energy in narrow bands around 8KHz, 16KHz, 22KHz, and 44KHz. 

\begin{figure}[!ht]
\begin{center}
    \includegraphics[width=1.1\linewidth]{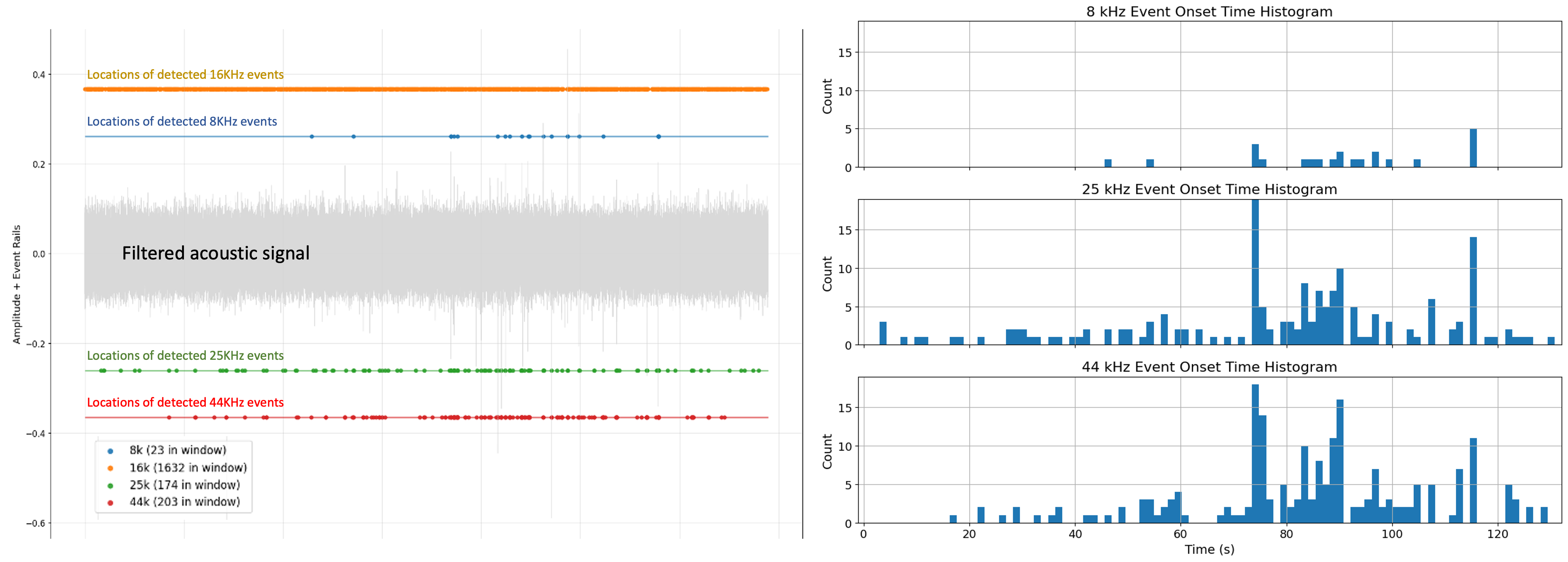}
    \caption{Visual representation of identified events in each of the frequency bands. On the left the gray central plot is the filtered waverform. The blue dots on the blue line indicate times where the onset of an AE event in the 8KHz band is detected. The orange, green, and red dots similarly represent the same for the 16KHz, 25KHz, and 44KHz events respectively. On the right these are shown in histogram form as well.} 
    \vspace{-0.2in}
    \label{fig:events_time}
\end{center}    
\end{figure}
Next we design targeted Morlet wavelets in narrow bands centered around the peak magnitude frequencies ($\Delta f = \pm 20\%$). Then using the \texttt{CWT} of the waveform we compute an "instantaneous band energy" metric of the form 
\begin{eqnarray*}
    E[t] & = & \sum_{\mbox{scales in band}} |\texttt{CWT}(\mbox{scale}, t) |^2
\end{eqnarray*}
This metric is then compared to a threshold to identify contiguous runs where the metric exceeds a band-specific threshold computed from the initial assumed dislocation-free time period. Using this approach we detect

\noindent
\begin{minipage}[t]{0.48\textwidth}
    \begin{itemize}
        \item 8KHz band: 23 events
        \item 16KHz band: 1,632 events
    \end{itemize}
\end{minipage}
\hfill
\begin{minipage}[t]{0.48\textwidth}
    \begin{itemize}
        \item 25KHz band: 174 events
        \item 44KHz band: 203 events
    \end{itemize}
\end{minipage}

Note that the events are not mutually exclusive -- there is overlap between these events. 
\begin{figure}[!ht]
\begin{center}
    \includegraphics[width=0.9\linewidth]{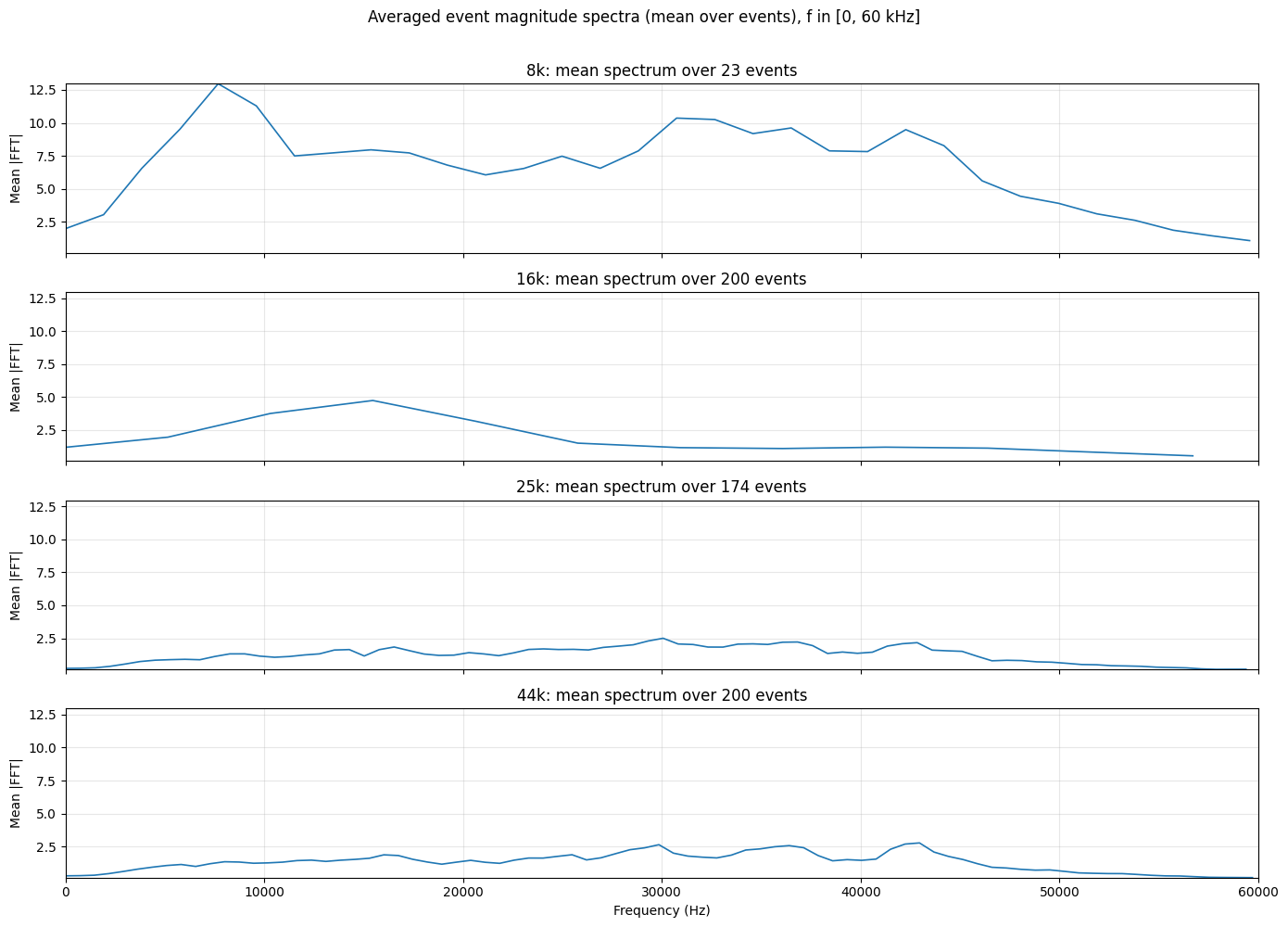}
    \caption{Average magnitude spectra for identified events in each of the frequency bands.} 
    \vspace{-0.2in}
    \label{fig:wavelet_band_spectra}
\end{center}    
\end{figure}

Figure (\ref{fig:events_time}) shows the locations of the onsets of the AE events relative to the signal and each other. The preponderance of events in the 16KHz band, including during the early stages of the signal that are expected to be dislocation-free, suggests it may not be a true artifact of the dislocations and is dropped in this analysis going forward. Having said that, the author will consult with his collaborators on this point to better understand if there may be a physical justification to retain them. The averaged magnitude spectra of the identified events in each of the frequency bands is shown in figure (\ref{fig:wavelet_band_spectra}).

Finally we compute a union of the events in the 8KHz, 25KHz, and 44KHz bands combining any overlapping events to obtain a final set of 266 unique AE events. These are shown for reference in the appendix in figure (\ref{fig:AE_events_over_time}). In addition for the classification task of the next section we extract a set of 'non-events' by randomly sampling portions of the signal that do not overlap with our detected events. 

\section{Physics-Based Validation of AE Event Detection}
\label{sec:physics}
\begin{figure}[!ht]
\begin{center}
    \includegraphics[width=1.1\linewidth]{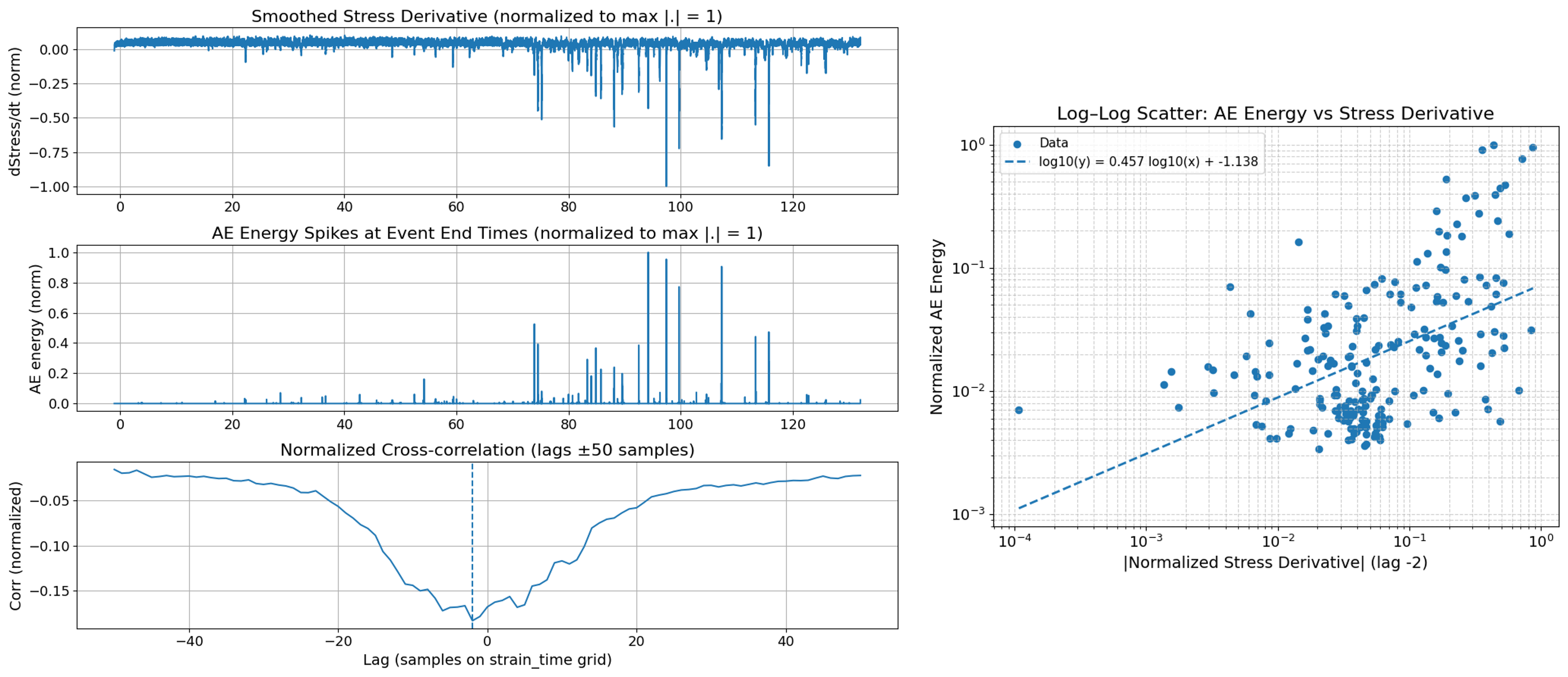}
    \caption{Correlation of the AE events with drops in the stress curve. The AE events line up well with the stress drops.} 
    \vspace{-0.2in}
    \label{fig:AEevents_vs_dstress}
\end{center}    
\end{figure}
AE emission events represent sudden releases of energy as the material internally adjusts to the increasing applied stress load stress. These adjustments manifest as sudden small dips in the stress curve. To test the validity of our AE event detection we compare the occurrence of the AE events to the derivative of the stress. We construct a signal comprised of AE events where each event is represented by a single sample of magnitude equal to the energy of the event (sum of square time values) located at the end of the event (detected onset of the event + the event duration). We can then correlate this with the stress derivative. This is shown in figure (\ref{fig:AEevents_vs_dstress}). The AE events line up well with the drops in stress and show an overall trend where the magnitude of the energy released also correlates with the size of the stress drop (with only a 2 sample lag).

\begin{figure}[!ht]
\begin{center}
    \includegraphics[width=0.8\linewidth]{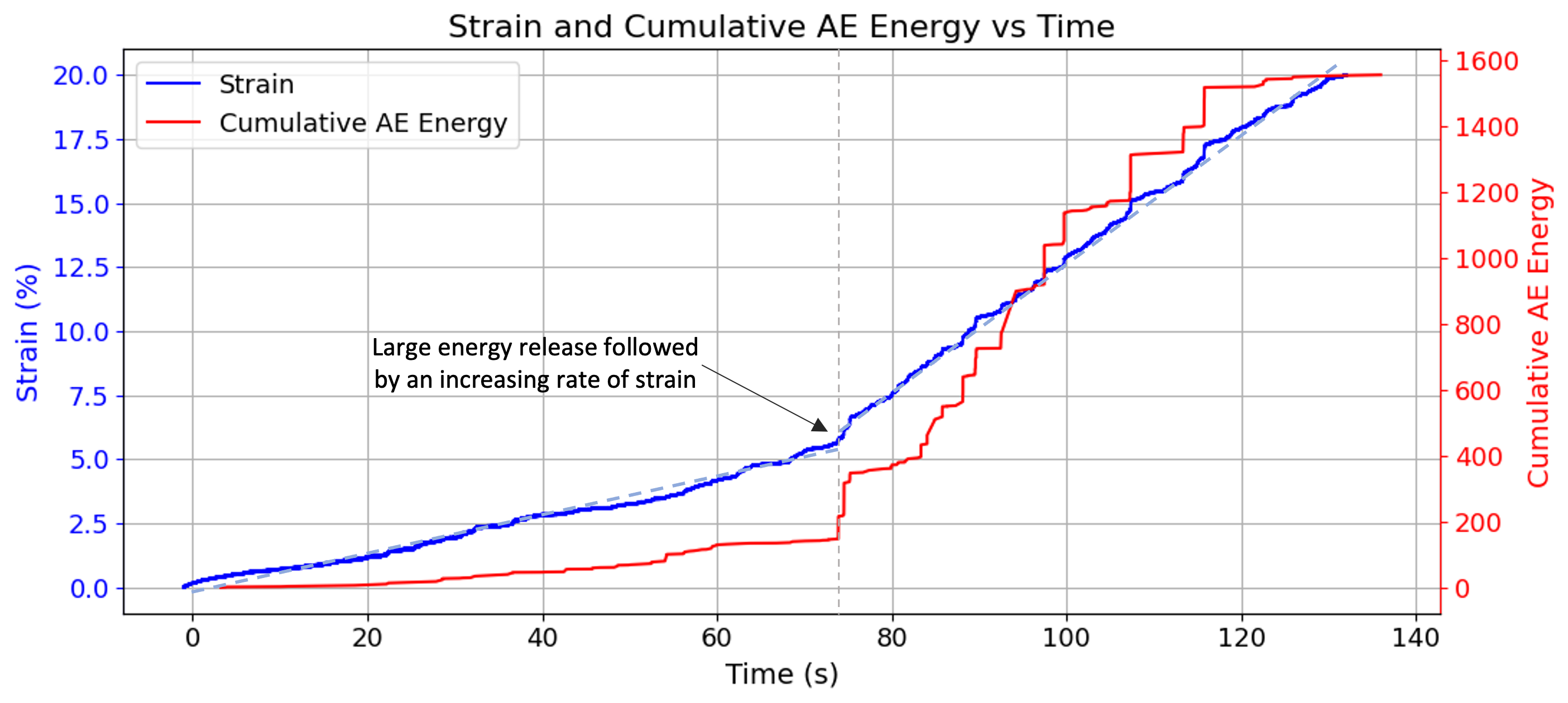}
    \caption{Strain and cumulative AE energy released over the experiment duration. We see the onset of increasing strain rate following a large AE event.} 
    \vspace{-0.2in}
    \label{fig:strain_AEenergy_annotate}
\end{center}    
\end{figure}
Figure (\ref{fig:strain_AEenergy_annotate}) shows the material strain and cumulative released AE energy of the events over time. We see that for the first half of the experiment small energy events and moderate strain coincide followed by a sudden large AE event followed by increasing strain rate.  

Together these plots provide validation that the AE event detection is consistent with expected physical interpretation.

\section{AE Event Classification and Archetype Identification}
\subsection{Naive KNN and SVM}
\begin{figure}[!ht]
\begin{center}
    \includegraphics[width=0.9\linewidth]{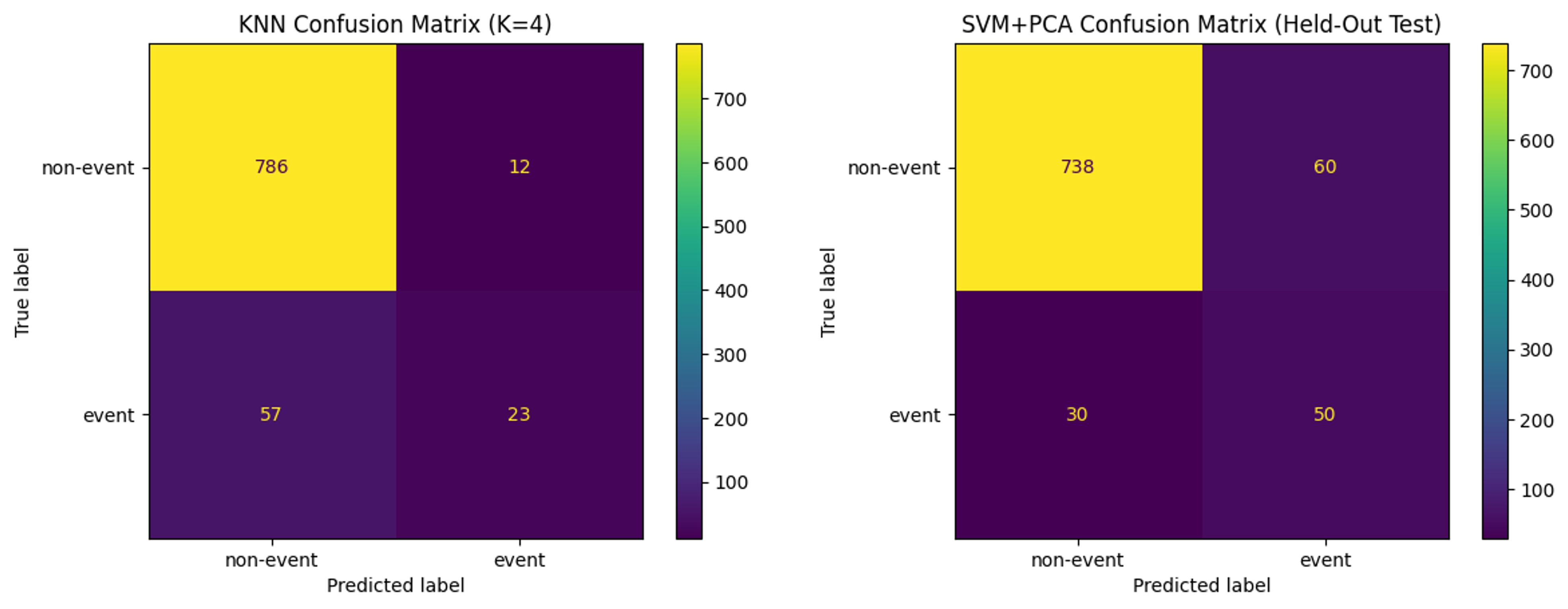}
    \caption{Confusion matrices on the test set for naive KNN (k=4) and SVM (with PCA) on the time domain signals.} 
    \vspace{-0.2in}
    \label{fig:KNN_SVM_Confusion}
\end{center}    
\end{figure}
As a baseline classifier we consider KNN and SVM-based classifiers on the time domain signals. Because snippets are variable length, make them comparable by resampling each snippet to a fixed length (e.g., 512 or 1024 samples) and then normalizing the amplitude using RMS. Further we standardize features to a zero mean / unit variance using StandardScaler. Note that our data is imbalanced. There are far more period of non-events than there are events. To represent this in the classification problem we use the identified events (260) and a sample of non-events that is 10 times its size (i.e., 2,600 non-events). 

KNN was tested for values of $K$ from 1 to 25, with 4 showing the best accuracy. SVM was tried with PCA dimension reduction (best result was 50 dimensions), different kernels (RBF was best). The resulting confusion matrices for both techniques is shown in figure (\ref{fig:KNN_SVM_Confusion}). We see that the class imbalance weighs on the results. KNN does a poor job of correctly classifying events (event recall = 29\%). SVM does a little better but still poorly (event recall = 63\%). 

\subsection{Time and Spectral Feature Classification}
Our next step is to compute a number of standard time and frequency domain features that have shown usefulness in seismic analysis (\cite{Ross18}, \cite{Anikiev23}) and investigate their ability to discriminate between events and non-events. We consider 12 features:

\noindent
\begin{minipage}[t]{0.48\textwidth}
    \begin{itemize}
        \item Peak amplitude: $\max |x(t)|$.
        \item RMS amplitude: $\sqrt{\frac{1}{N} \sum x^2}$.
        \item Signal energy: $\sum x(t)^2$.
        \item Crest factor: $\frac{\max |x|}{\mbox{RMS}}$ (impulsiveness).
        \item Kurtosis
        \item Skewness (asymmetry)
    \end{itemize}
\end{minipage}
\hfill
\begin{minipage}[t]{0.48\textwidth}
    \begin{itemize}
        \item Zero crossing rate (no. sign changes).
        \item Rise time: $t_{\mbox{peak}} - t_{\mbox{onset}}$.
        \item Decay time of signal envelope.
        \item Spectral centroid: $f_c = \frac{\sum f S(f)}{\sum S(f)}$.
        \item Spectral bandwidth
        \item Spectral entropy: $-\sum p(f) \log p(f)$.
    \end{itemize}
\end{minipage}

Using these features we again train a KNN classifier and SVM classifier and achieve much improved results on the test set:

{\footnotesize
\begin{verbatim}
Best KNN k: 3  test acc: 0.985
Confusion Matrix
 [[531   1]
 [  8  45]]    

Best SVM params: {'svm__C': 1, 'svm__gamma': 0.01, 'svm__kernel': 'rbf'}  test acc: 0.991
Confusion Matrix (SVM):
 [[527   5]
 [  0  53]] 
\end{verbatim}}

\begin{figure}[!ht]
\begin{center}
    \includegraphics[width=0.9\linewidth]{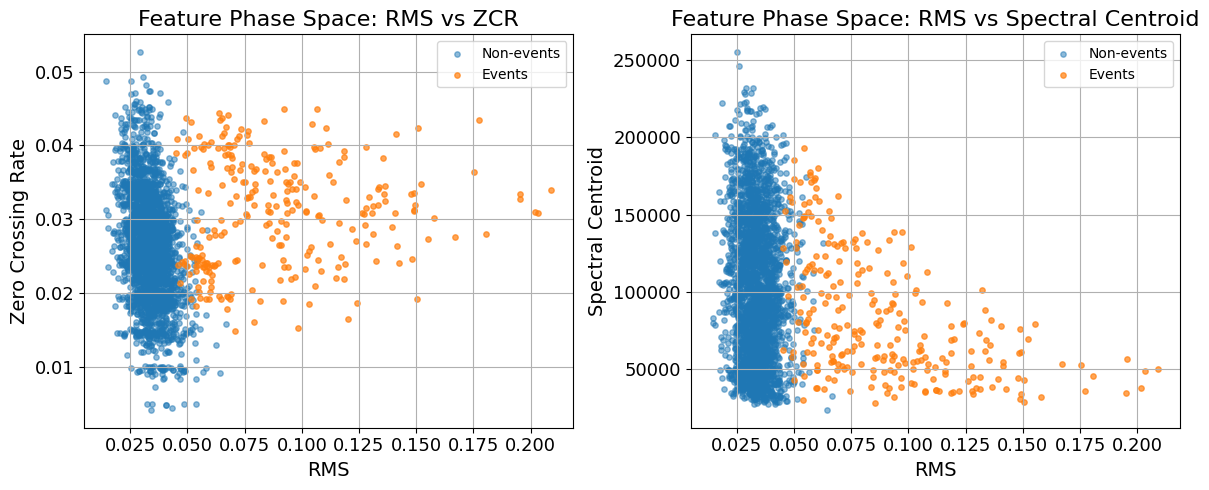}
    \caption{Scatter plot of events and non-events in the 2D feature spaces of RMS vs ZCR an RMS vs spectral centroid. The plots show good clustering and separation of events and non-events along these features.} 
    \vspace{-0.2in}
    \label{fig:2d_feature_phase_plot}
\end{center}    
\end{figure}
To understand the importance of each of these parameters we perform a permutation importance on the best SVM with a scoring metric of accuracy and find that signal strength (RMS), zero crossing rate,  and spectral centroid have the biggest impacts on accuracy. Figure (\ref{fig:2d_feature_phase_plot}) shows graphically that these features do well in separating the events and non-events. These results are unsurprising and replicate those of seismic events: AE events produce sudden strong impulsive bursts of high energy at certain characteristic frequencies. 

\subsection{Event Archetype Identification via Clustering}
\begin{figure}[!ht]
\begin{center}
    \includegraphics[width=1.1\linewidth]{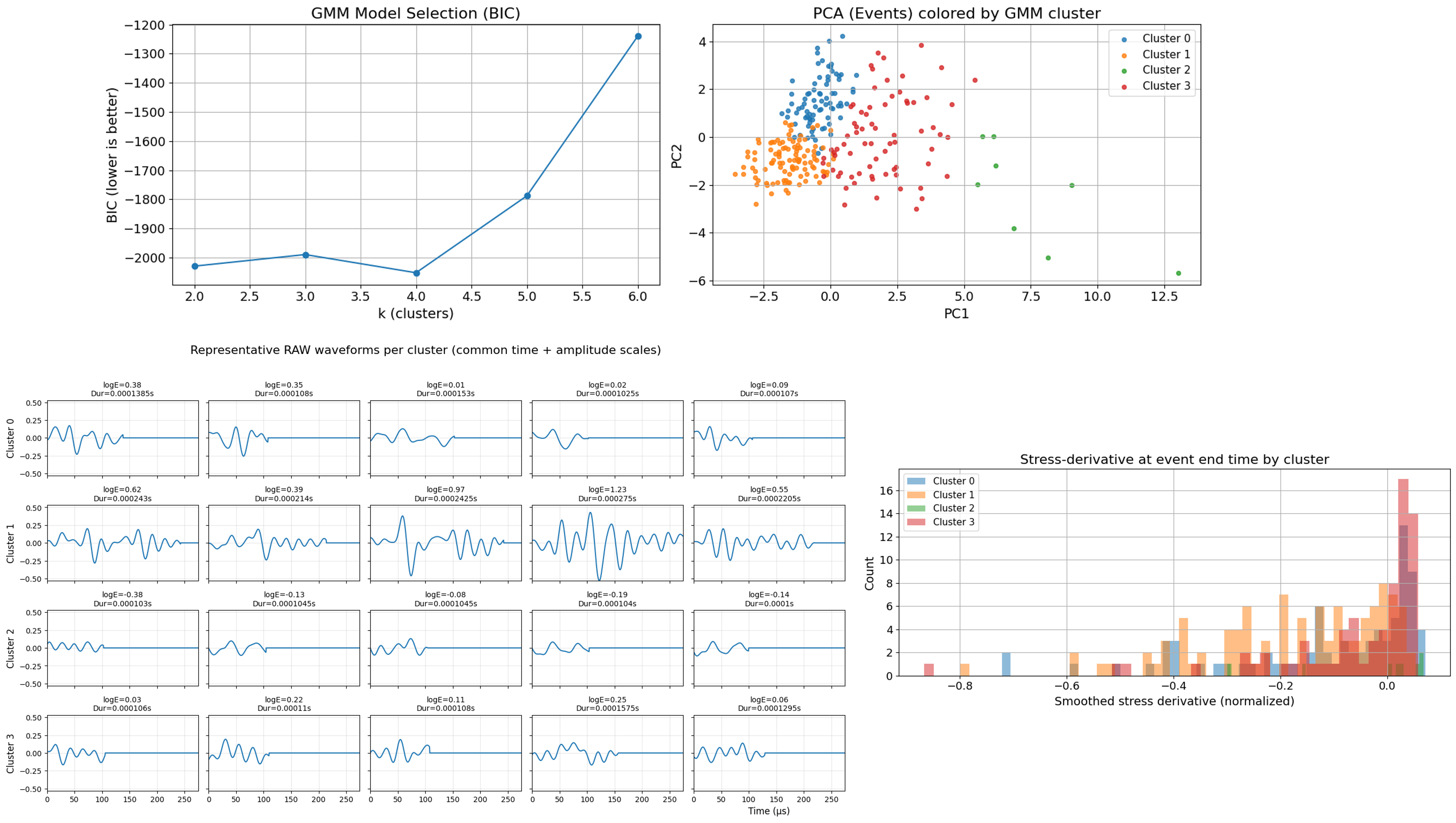}
    \caption{Identified clusters using a variety of time and frequency features. In the top figures we identify 4 clusters via a Gaussian Mixture Model and visualize them in 2 dimensions. In the lower left we show sample waveforms in each cluster. In the lower right we overlay the histogram of cluster occurrences over time.} 
    \vspace{-0.2in}
    \label{fig:clustering_combo}
\end{center}    
\end{figure}
Using the labeled dataset Next we perform an analysis to identify the most impactful signal features in distinguishing AE events from non-events. We start by extracting a compact feature vector from each event and non-event waveform that emphasizes signal morphology and time–frequency distribution, including: event duration, envelope rise and decay fractions computed from the Hilbert envelope, kurtosis of the RMS-normalized waveform (impulsiveness), a zero-crossing–rate-based frequency proxy converted to crossings/second using the known sampling rate (2 MHz), and FFT-based spectral descriptors (centroid, bandwidth in kHz, and spectral entropy). Additionally the vector is augmented with wavelet energy fractions across DWT sub-bands plus a wavelet entropy, which can help separate broadband “bursty” events from lower-frequency or more structured emissions. All features are then standardized (z-scored) so no single scale dominates, reduced to 2D with PCA for visualization, and clustered with a Gaussian Mixture Model (GMM). The number of clusters is selected by sweeping $K = 2, \ldots, 6$ and choosing the model with the lowest BIC (a likelihood-with-complexity-penalty criterion). Figure (\ref{fig:clustering_combo}) visualizes: (i) the BIC curve, (ii) the PCA scatter colored by cluster,  (iii) plots representative raw waveforms per cluster (chosen as the events closest to each cluster mean in standardized feature space) using a common time and amplitude scale, and (iv) a histogram mapping each event’s end time to the nearest value of a provided smoothed stress-derivative signal (relates clusters to mechanics).

Note in our clustering we incorporate wavelet-domain features to capture how signal energy is distributed across time–frequency scales, which is particularly useful for transient signals like acoustic emissions. Using the discrete wavelet transform (DWT) with a Daubechies-4 (DB4) mother wavelet, the waveform is decomposed into multiresolution components at level 3, producing approximation and detail coefficients corresponding to progressively lower frequency bands. We compute the fraction of total signal energy contained in each sub-band (D1, D2, D3, and A3), along with a wavelet entropy, which measures how concentrated or spread the energy is across those scales. These features help distinguish between events that are dominated by high-frequency, impulsive energy (often associated with brittle microcracking or rapid slip) and those with lower-frequency or more distributed energy (which may correspond to frictional sliding, rubbing, or slower fracture processes). The DB4 wavelet is commonly used in acoustic emission and transient signal analysis because its compact support and moderate number of vanishing moments allow it to capture sharp transients while still representing oscillatory structure effectively \cite{Hamstad02}. In practice, DB4 provides a good compromise between time localization (detecting short bursts) and frequency discrimination, making it well suited for AE signals that often consist of short, broadband bursts with rapidly changing frequency content.

\begin{table}[ht]
\centering
\tiny
\setlength{\tabcolsep}{4pt} 
\begin{tabular}{ccccccccccc}
\toprule
Cluster & Count & RMS & Peak$_{\mathrm{abs}}$ & Energy & Kurtosis & ZCR & Spec. Centroid & Spec. Entropy & Median Duration & Median $d\sigma/dt$ \\
 &  &  &  &  &  & (cross/sample) & (KHz) &  & ($\mu$s) &  \\
\midrule
1 & 105 & 0.112  & 0.340 & 4.995  & 4.166 & 0.032 & 51.5  & 2.75  & 207   & -0.134 \\
0 & 80  & 0.0750 & 0.171 & 1.585  & 2.654 & 0.025 & 106.1  & 3.05 & 116   & -0.0277 \\
3 & 73  & 0.0639 & 0.138 & 0.915 & 2.395 & 0.034 & 86.8  & 2.70 & 110   & 0.00816 \\
2 & 8   & 0.0569 & 0.109 & 0.693 & 2.004 & 0.022 & 61.1  & 2.40 & 105 & 0.0235 \\
\bottomrule
\end{tabular}
\vspace{0.1in}
\caption{Cluster summary (sorted by number of events). Values are medians within each cluster. ZCR is reported in crossings/sample, spectral centroid in KHz, and duration in $\mu$s.}
\label{tab:cluster_summary}
\end{table}

Table (\ref{tab:cluster_summary}) shows a summary of representative key features in each cluster. From the table we can describe the clusters as follows (in order of most frequently occurring):
\begin{enumerate}
    \item Cluster 1 (40\% of events) — High-energy impulsive events: This cluster contains the highest-amplitude and highest-energy signals, with the largest kurtosis and longest median duration, indicating strongly impulsive AE bursts. Their relatively low spectral centroid suggests broadband events dominated by lower frequencies, consistent with larger fracture or slip processes.

    \item Cluster 0 (30\% of events) — High-frequency distributed events: These events have moderate amplitude and energy but the highest spectral centroid and spectral entropy, indicating relatively high-frequency and spectrally distributed signals. Their shorter duration and lower kurtosis compared with Cluster 1 suggest smaller-scale, less impulsive acoustic emissions.

    \item Cluster 3 (27\% of events) — Rapid oscillatory bursts: Cluster 3 is characterized by the highest zero-crossing rate and intermediate spectral centroid, indicating signals with rapid oscillatory content. These events have relatively low energy and short durations, consistent with small, fast transient emissions.

    \item Cluster 2 (3\% of events) — Low-energy compact events: This cluster contains the lowest-amplitude and lowest-energy events with the smallest kurtosis, indicating relatively smooth or weak AE signals. Their short duration and modest frequency content suggest minor microstructural adjustments or small-scale precursory activity.
\end{enumerate}

\section{Strain Regime Change Prediction}
\begin{figure}[!ht]
\begin{center}
    \includegraphics[width=0.7\linewidth]{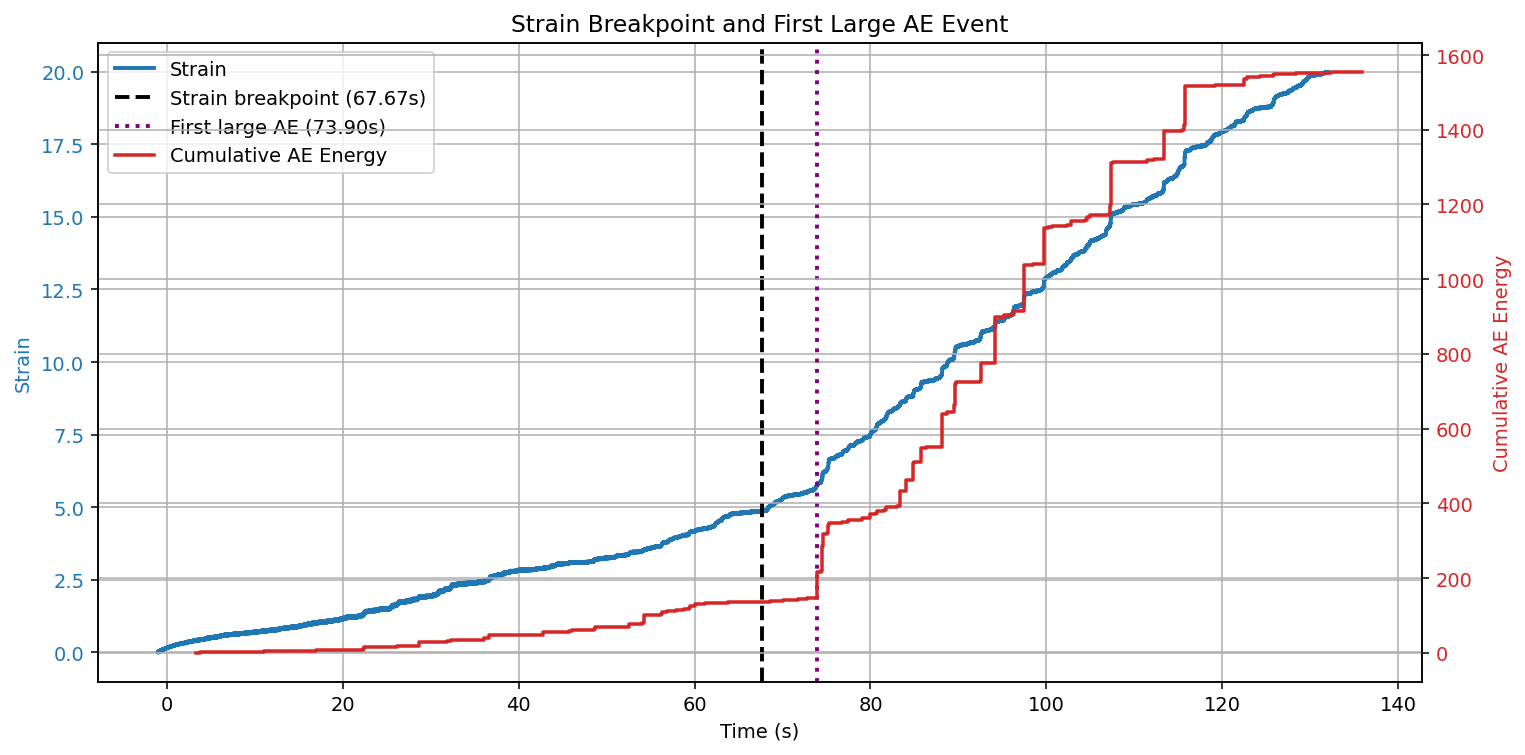}
    \caption{Identified clusters using a variety of time and frequency features. In the top figures we identify 4 clusters via a Gaussian Mixture Model and visualize them in 2 dimensions. In the lower left we show sample waveforms in each cluster. In the lower right we overlay the his} 
    \vspace{-0.2in}
    \label{fig:strain_breakpoint}
\end{center}    
\end{figure}
In this section we present some preliminary results investigating the predictive potential of our event archetypes. Figure (\ref{fig:strain_breakpoint}) shows the strain curve again with cumulative AE energy. As was mentioned in section \ref{sec:physics} the strain curve exhibits a clear change in slope (strain-rate increases) and AE activity jumps. We therefore ask the question whether we can detect the onset of this deformation regime change. To begin, we fit a best 2 line curve to the strain curve to detect the onset of regime change which occurs around 67 seconds into the experiment. We use this point to delineate the point of regime change. 

\begin{figure}[!ht]
\begin{center}
    \includegraphics[width=0.75\linewidth]{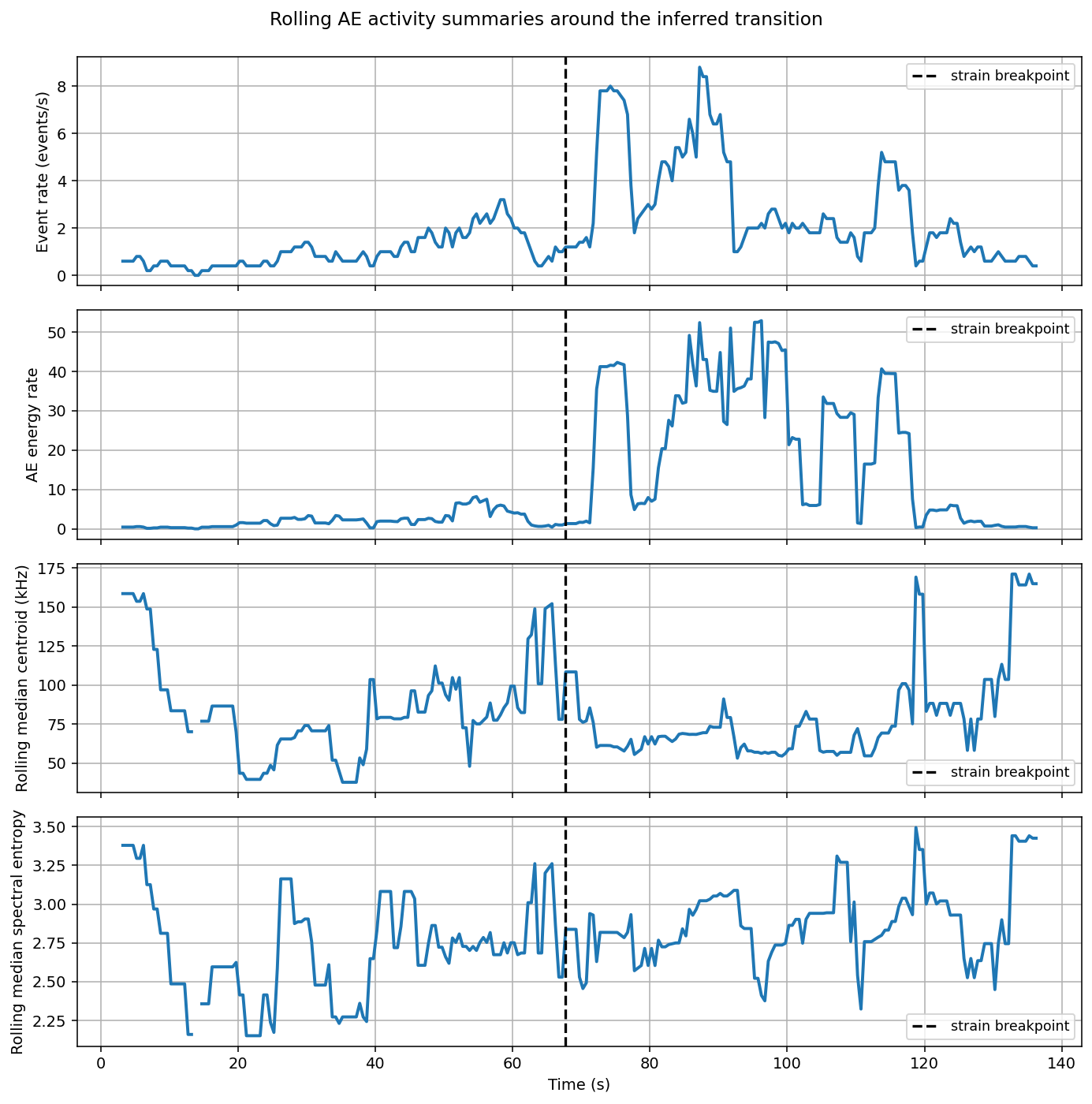}
    \caption{Identified clusters using a variety of time and frequency features. In the top figures we identify 4 clusters via a Gaussian Mixture Model and visualize them in 2 dimensions. In the lower left we show sample waveforms in each cluster. In the lower right we overlay the his} 
    \vspace{-0.2in}
    \label{fig:before_after}
\end{center}    
\end{figure}
Figure (\ref{fig:before_after}) shows the key feature characteristics of the AE events before and after the regime change. We summarize our observations of the before and after in table (\ref{tab:ae_transition_features}). 
\begin{table}[ht]
    \centering
    \footnotesize
    \begin{tabular}{lll}
        \toprule
        Feature & Before Transition & After Transition \\
        \midrule
        Event rate & Moderate & Pronounced increase \\
        AE energy rate & Small & Large increase \\
        Spectral centroid & More high-frequency activity & Lower frequencies dominate initial transition \\
        Spectral entropy & Some spectral concentration of energy & More broadband activity \\
        \bottomrule
    \end{tabular}
    \caption{Qualitative comparison of acoustic emission (AE) characteristics before and after the onset of accelerated strain.}
\label{tab:ae_transition_features}
\end{table}

\begin{figure}[!ht]
\begin{center}
    \includegraphics[width=0.7\linewidth]{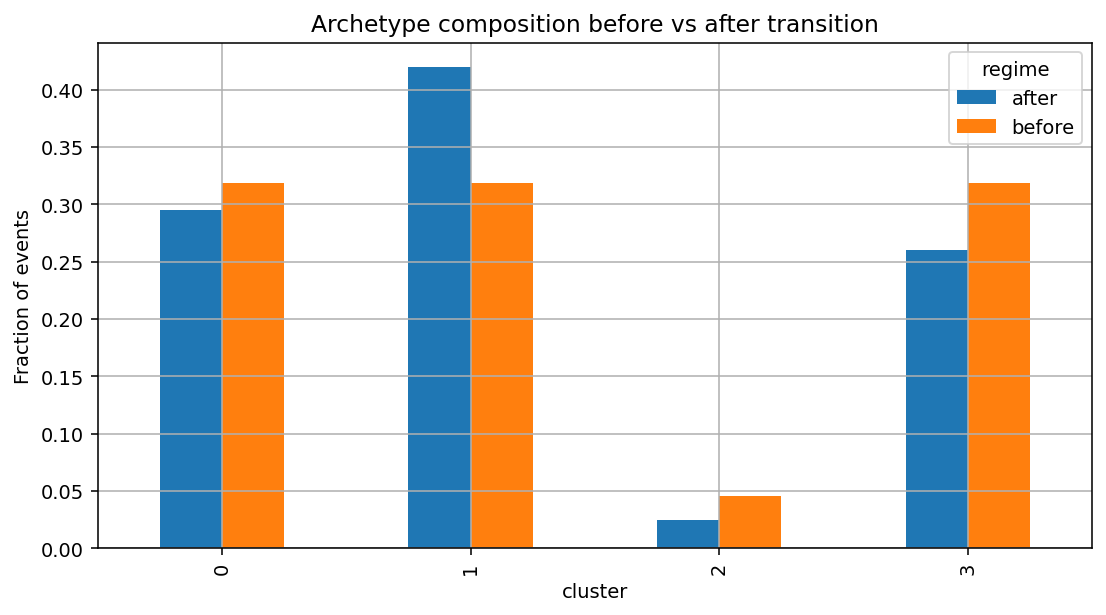}
    \caption{Frequency of clusters before and after the strain regime change.} 
    \vspace{-0.2in}
    \label{fig:cluster_before_after}
\end{center}    
\end{figure}
Figure (\ref{fig:cluster_before_after}) shows the frequency of the archetypes before and after the strain regime change. We see that after the change cluster 1 shows a significant increase in prevalence, which makes sense as these are the high-energy impulsive events. We also note a higher prevalence of clusters 2 and 3 before the regime change which are the higher-frequency oscillatory events. 

\section{Conclusion and Next Steps}
The analysis demonstrates that wavelet-based detection combined with feature-based machine learning provides a physically meaningful and effective framework for identifying and characterizing acoustic emission (AE) events during plastic deformation. By detecting events in specific frequency bands using Morlet wavelets and validating them against stress-drop dynamics, the study shows that the detected AE events correlate strongly with underlying mechanical processes in the material. Clustering of these events reveals four distinct waveform archetypes, likely corresponding to different scales or mechanisms of plastic activity, ranging from high-energy impulsive bursts to smaller high-frequency transient signals. Further, the features of AE events show differences before and after distinctive changes in the strain curve, as do the frequency of archetypes.

The next steps should focus on more in-depth predictive modeling of plastic deformation dynamics. This could include
\begin{enumerate}

    \item analyzing the temporal evolution of AE spectral features to identify precursors to large dislocation avalanches, 
    
    \item modeling AE activity as sequential processes using time-series or sequence models to capture correlations between events, and 
    
    \item applying additional change-point detection or regime identification to determine when the material transitions between deformation modes. 
\end{enumerate}
Ultimately, combining these approaches with physics-informed constraints could enable forecasting of large plastic events or strain-rate changes from early AE activity, providing a path toward predictive acoustic monitoring of material deformation.

\section{Code}
All code for this report is located here:
\begin{verbatim}
https://github.com/kelawady/Acoustic_Emissions
\end{verbatim}

\bibliographystyle{acl_natbib}
\bibliography{references}

\newpage
\begin{center}
Appendix
\end{center}
\vspace{-0.25in}
\begin{figure}[!ht]
\begin{center}
    \includegraphics[width=0.79\linewidth]{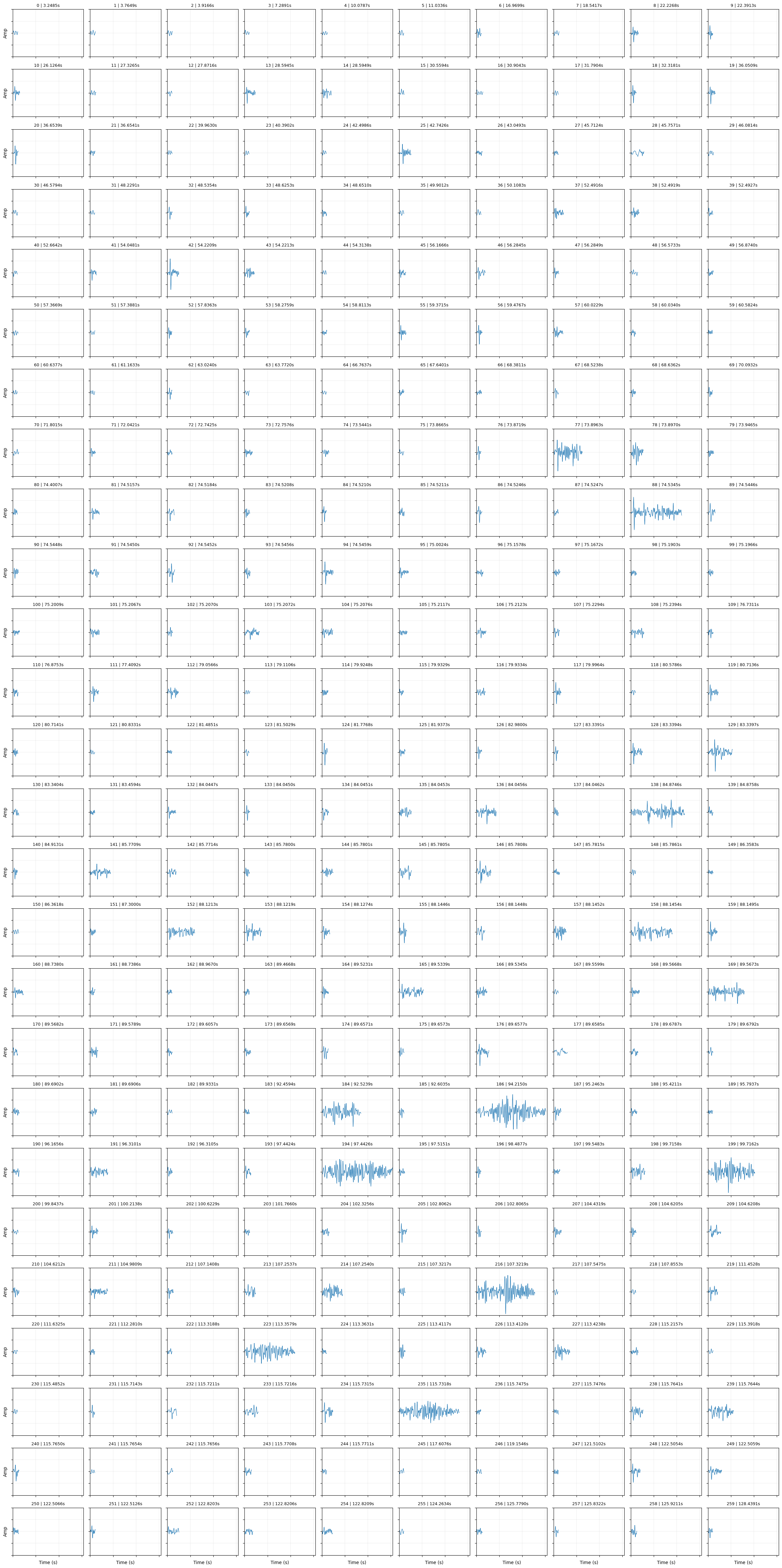}
    \caption{AE events over time as they occur.} 
    \vspace{-0.35in}
    \label{fig:AE_events_over_time}
\end{center}    
\end{figure}

\end{document}